\documentclass{article}

\usepackage{arxiv}

\usepackage[utf8]{inputenc} 
\usepackage[T1]{fontenc}    
\usepackage{hyperref}       
\usepackage{url}            
\usepackage{booktabs}       
\usepackage{nicefrac}       
\usepackage{microtype}      
\usepackage{lipsum}
\usepackage{amsmath,amssymb,amsfonts}
\usepackage{algorithmic}
\usepackage{graphicx}
\usepackage{textcomp}
\usepackage{xcolor}
\usepackage{hyperref}

\title{Causal Influences Decouple From Their Underlying Network Structure In Echo State Networks}

\author{
 Kayson Fakhar \\
  Institute of Computational Neuroscience\\
  University Medical Center Hamburg-Eppendorf (UKE)\\
  Hamburg, Germany \\
  \texttt{k.fakhar@uke.de} \\
   \And
 Fatemeh Hadaeghi \\
  Institute of Computational Neuroscience\\
  University Medical Center Hamburg-Eppendorf (UKE)\\
  Hamburg, Germany \\
  \texttt{f.hadaeghi@uke.de} \\
  \And
 Claus C. Hilgetag \\
  Institute of Computational Neuroscience\\
  University Medical Center Hamburg-Eppendorf (UKE)\\
  Hamburg, Germany \\
  Department of Health Sciences\\ 
  Boston University\\ 
  Boston, MA, USA\\
  \texttt{ c.hilgetag@uke.de} \\
}
\date{ }
\begin{document}
\maketitle
\begin{abstract}
Echo State Networks (ESN) are versatile recurrent neural network models in which the hidden layer remains unaltered during training. Interactions among nodes of this static backbone (the network structure) produce diverse representations (\textit{i.e.,} network dynamics) of the given stimuli that are harnessed by a read-out mechanism to perform computations needed for solving a given task (\textit{i.e.,}  behavior). Moreover, ESNs are accessible models of neuronal circuits, since they are relatively inexpensive to train. Therefore, ESNs have become attractive for neuroscientists studying the relationship between neural structure, function, and behavior. For instance, it is not yet clear how distinctive connectivity patterns of brain networks (structure) support effective interactions among their nodes (dynamics) and how these patterns of interactions give rise to computation (behavior).

To address this question, we employed an ESN with a biologically inspired structure and used a systematic multi-site lesioning framework to quantify the causal contribution of each node to the network's output, thus providing a causal link between network structure and behavior. We then focused on the structure-function relationship and decomposed the causal influence of each node on all other nodes, using the same lesioning framework.

We found that nodes in a properly engineered ESN interact with each other largely irrespective of the network's underlying structure. However, in a network with the same topology where the ESN's leakage rate is non-optimal and the dynamics are diminished, the underlying connectivity patterns determine the node interactions. 

Our results suggest that causal structure-function relations in ESNs can be decomposed into two components, \textit{direct} and \textit{indirect} interactions. The former are based on influences relying on structural connections. The latter describe the effective communication between any two nodes through other intermediate nodes. These widely distributed indirect interactions may crucially contribute to the efficient performance of ESNs.
\end{abstract}


\section{Introduction}\label{introduction}

One of the fundamental goals of neuroscience is to understand neuronal mechanisms underlying cognition and behavior \cite{Adolphs2016-vq}. This goal has remained challenging because the brain consists of a highly complex web of interconnected units (connectome) that interact to process, transform, and transfer information (dynamics/function), in order to achieve a performance outcome (cognition and behavior) \cite{Vyas2020-tq}. Previous investigations through mathematical models of neuronal dynamics showed that, in such networks, nodes not only influence their immediate neighbors, but also affect spatially and topologically remote units, such that local perturbations generate global effects \cite{Alstott2009-ka}. One reason for this network-wide impact is that nodes influence each other through routes other than direct structural connections, for example, via intermediate nodes \cite{Avena-Koenigsberger2018-qi}, such that lesioning a node may also perturb nodes that are not directly connected to the lesioned site \cite{Young2000-cz}. However, since many mathematical models of the brain lack a behavioral output \cite{Breakspear2017-yf} and an extensive perturbation of real brain networks is not possible, the relationship between structure, function, and behavior
of neural networks is not yet fully understood.

Artificial neural network (ANN) models have found a unique place in the toolbox of computational neuroscientists during the past decade. ANNs have made studying the relations between structure, function, and behavior increasingly more accessible, since their ground-truth connectome is known and parametrizable \cite{Pulvermuller2021-fo,Goulas2021-bg}. Among the large family of ANNs, reservoir computing models \cite{Jaeger2004-qb,Maass2002-hp} are particularly suitable for investigating the relationship between structure, function, and behavior. The reasons are (i) the underlying network structure remains unchanged during the training process, (ii) with an appropriate selection of the hyper-parameters, a rich repertoire of heterogeneous representations is generated by the reservoir and (iii) the harvested representations (i.e., function) are used to produce quantifiable behaviors, such as solving a short-term memory task \cite{Damicelli2021-lr,Su2020-xz}.

To investigate how nodes communicate with each other across a network and contribute to the system's behavior, we trained a compact ESN with a biologically plausible connectome to achieve a task of time-series prediction. Then, we systematically perturbed the hidden layer of the ESN and ranked its nodes based on their contributions to the network's output signal. In this way, we were able to causally map nodes to the behavior, such that perturbing the most critical units was functionally detrimental, by pushing the system towards pathological dynamics leading to distorted target predictions. Afterwards, we applied the same framework to quantify the effect of lesioning each node on the produced dynamics of other nodes in the hidden layer. Here, for each node, we derived the rankings of other nodes, such that removing the most influential nodes caused severe disruption in the target node's dynamics. Iterating over all nodes resulted in a `causal influence matrix' representing the causal interactions existing among nodes. This causal influence matrix included two additive components, namely \textit{direct} and \textit{indirect} influences, where the former was characterized by the direct structural connections, and the latter represented the interaction between unconnected nodes  requiring paths via other intermediate nodes. For comparison, we then performed the same analysis on a control network with the same topology where the ESN's leakage rate was set to a value 100 times smaller than the optimal value. This significant increase in the memory of each node led to a less diverse repertoire of state representations and a mediocre task performance (Mean Squared Error; MSE = 0.055 instead of 0.0049 for the optimally configured ESN). 
Our simulations showed that, unlike in the well-engineered ESN, in this poorly designed control ESN, causal influences were more related to the connectome of direct structural connections and indirect influences were less significant. Furthermore, we observed the same phenomenon, i.e. a closer relation to the structural connectome, in a simpler cellular automaton-based information propagation model using an identical structural connectome. In this experiment, the state of each node was determined by a susceptible-excited-refractory (SER) cyclic mechanism.
Finally, we ranked the nodes of the intact ESN according to their influence on each other and compared this ranking with the ranking derived from the nodes' contributions to the network behavior. A significant relationship between these two rankings was observed which indicates that perturbing the critical nodes destabilizes the internal dynamics which, in turn, results in disrupted behavior.

\section{Materials and Methods}\label{Materials}
\subsection{Network Topology, Models, and Their Dynamics}\label{Topology and Models}
\begin{figure}[hbtp]
\centerline{\includegraphics[scale=1]{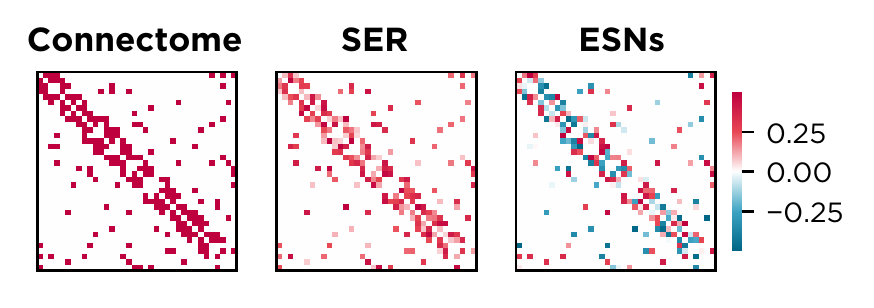}}
\caption{\textbf{Structural connectivity matrices:} A small-world network was used as the structural backbone of both models. The wights were sampled from uniform distributions over [0.1, 1] and [-0.5, 0.5] in SER and ESN models, respectively.}

\label{fig1}
\end{figure}

For the connectome, in both SER and ESN setups, we exploited the `small-world' architecture \cite{Bullmore2009-oh,Watts1998-sr}, one of the most frequently observed connectivity patterns in complex systems, including the brain. The reason why this motif is abundant is argued to be a trade-off between the topological efficiency of direct connections and the material cost of them when it comes to spatially distant nodes. In small-world networks, neighboring nodes are densely connected, forming local communities, and the number of nodes traversed from and to any two nodes is small due to a few long-range connections that bridge these local clusters. Hence, this topology is more efficient than a lattice network and less wire-expensive than a randomly connected pool of nodes \cite{Bullmore2012-xu}. In this study, the connectome consisted of 36 nodes and 216 connections, where each node was connected to its six nearest neighbors, forming ring-like communities. The probability of connecting to spatially farther communities was set to $p=0.4$, which resulted in a handful of long-range connections. The connection weights were randomly sampled from a uniform distribution with real values between [-0.5, 0.5] for ESNs and [0.1, 1] for the SER network (Fig.\ref{fig1}).

\subsubsection{Echo State Network and Its Task}\label{ESN}

We exploited a leaky ESN model \cite{Jaeger2007-yj} with the network state update equation given in Eq.~\ref{eq: eq1}:
\begin{equation}
\label{eq: eq1}
\begin{split}
\mathbf{x}(t+1) & = (1- a \eta)\mathbf{x}(t) + \eta f (\mathbf{W}^{\text{in}}\mathbf{u}(t+1) + \mathbf{W}\mathbf{x}(t)+\mathbf{W}^{\text{fb}}\mathbf{y}(t)),
\end{split}
\end{equation}
where $\mathbf{u}(t) \in \mathbb{R}^{N_{u}}$ denotes the external input, $\mathbf{x}(t) \in \mathbb{R}^{N_{x}}$ is the time-dependent $N_{x}$-dimensional reservoir state (\textit{i.e,} a reservoir with $N_{x}$ units), $\mathbf{W}\in\mathbb{R}^{N_x\times N_x}$, $\mathbf{W}^{in}\in\mathbb{R}^{N_x\times N_u}$, and  $\mathbf{W}^{\text{fb}} \in\mathbb{R}^{N_x\times N_y}$ indicate internal, input-to-reservoir, and feedback (output-to-reservoir) weight matrices, respectively. In this equation, $a>0$ and $\eta>0$ are reservoir neurons' leakage rate and time constant. The leakage rates for the intact and control ESNs were set to 0.1 and 0.001, respectively. The internal states were updated via the non-linear function, $f$, and the output $\mathbf{y}(t)\in \mathbb{R}^{N_{y}}$ is obtained by
\begin{equation}
\label{eq: eq2}
\mathbf{y}(t)=g\left(\mathbf{W}^{\text {out }}[\mathbf{x}(t) ; \mathbf{u}(t)]\right)
\end{equation}
where the vertical vector concatenation $[.;.]$ is applied to calculate the extended system state, $\mathbf{z}(t) = [\mathbf{x}(t); \mathbf{u}(t)]$. In Eq.~\ref{eq: eq2}, $g$ denotes an output activation function and $\mathbf{W}^{out}\in\mathbb{R}^{N_y\times(N_x+N_u)}$ is the trainable readout weight matrix.

Since the network is compact, we used a non-trivial but manageable task, \textit{i.e,} prediction of a chaotic time-series generated by the Mackey-Glass time-delay differential equation:
\begin{equation}
\frac{dr}{dt} = \frac{0.2r(t-\tau)}{(1+r(t-\tau)^{10}-0.1r(t))}
\end{equation}
Following the literature, we set the time delay $\tau = 17$ and simulated the equation using the \textit{dde23} solver for delay differential equations from the commercial toolbox MATLAB. Fig.\ref{fig3}.A depicts 500 steps of the obtained data. In our ESN experiments, the length of training and test sequences were 2500 and 500, respectively. The read-out weights were computed via the pseudoinverse method to minimize the least square error between the target outputs and the already harvested extended system state, $\mathbf{z}(t)$. We optimized the spectral radius of the internal weight matrix, $\rho$, and the neurons' leakage rate, $a$, by performing extensive grid-search on 50 independently created 36-unit ESNs with the same small-world topology. We set $\eta = 1.0$ and used the hyperbolic tangent function, $f(.)=\tanh(.)$, for the reservoir neurons and an identity function for the readout node.

\begin{figure*}[ht]
\centering
\includegraphics[scale=0.9]{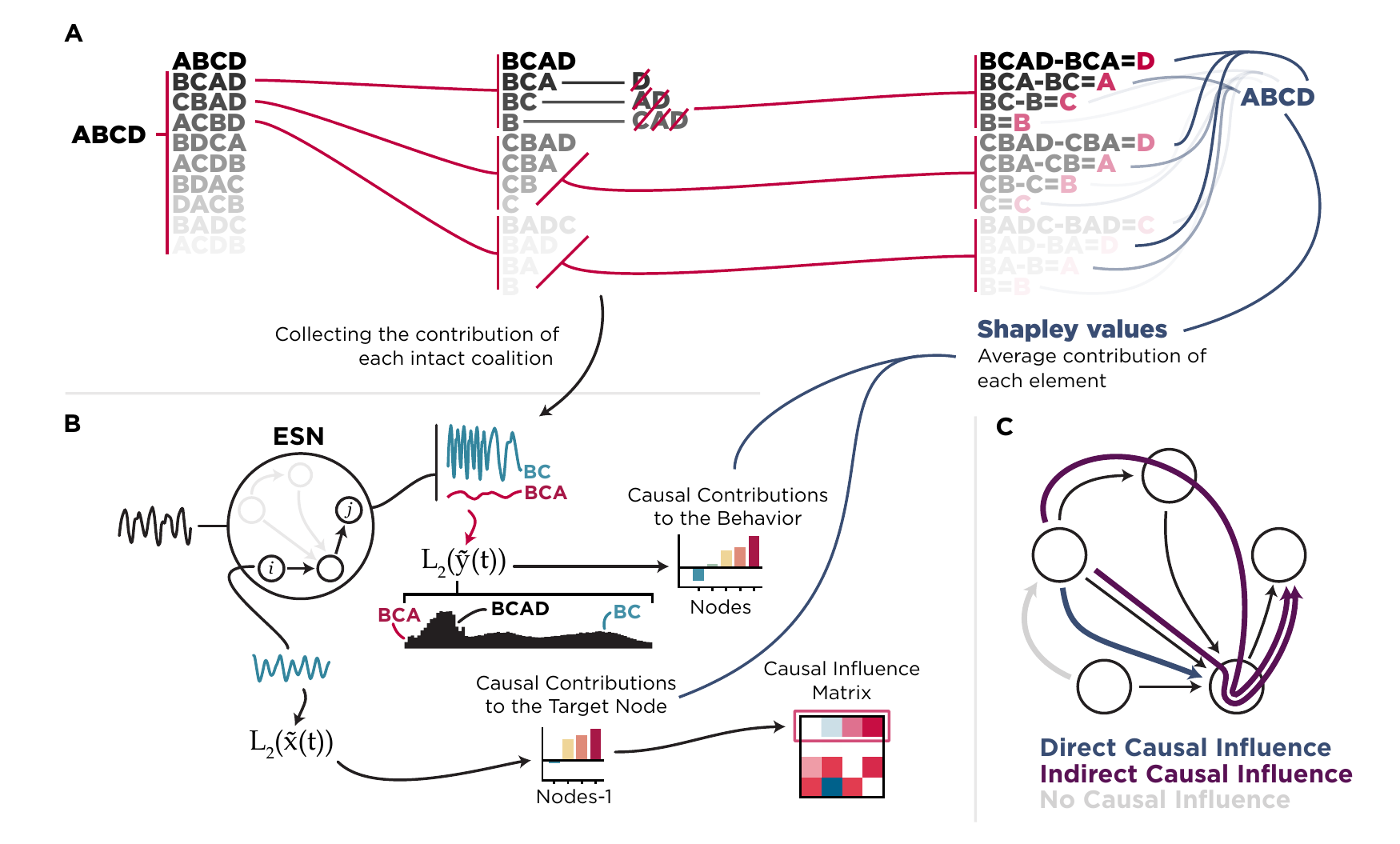}
\caption{\textbf{Schematics of the methods: A)} The Multi-perturbation Shapley value Analysis (MSA) was used to systematically lesion nodes throughout this work. MSA starts by first permuting the given set of players to form the `permutation space'. The permutation space then dictates which combinations should be perturbed, and a simple difference between two combinations, once with and once without an arbitrary element, quantifies the contribution of that element to that specific combination. Averaged over all permutations provides the marginal contribution of that player to the grand coalition, derived from a dataset of multi-site lesions. \textbf{B)} To decompose the causal influences of nodes in the networks, we used MSA on each node such that for each node, the rest of the network formed the player-set. The contribution is then defined as the impact of each node on the energy of the target node, $L_2(\tilde{x}(t))$. Nodes positively contributing to the energy level of other units were those that, when removed, averaged over all coalitions, resulted in a smaller $L_2(\tilde{x}(t))$, and vice versa for negatively contributing nodes. In this illustration, removing element D from the coalition BCAD resulted in a signal with lower energy and, consequently, smaller  $L_2(\tilde{x}(t))$. However, removing the pair AD had the opposite effect by over-activating the network, which is summarized with a large  $L_2(\tilde{x}(t))$. See Fig\ref{fig4} for the actual histogram of $L_2(\tilde{x}(t))$. \textbf{C)} Causal influence decomposition captures both direct and indirect causal influences. If there is a structural connection between two nodes, their corresponding causal influence is assumed to be direct. Suppose the nodes share no structural connections and a perturbation in one node resulted in an alteration of $L_2(\tilde{x}(t))$ in the other. In that case, the effect must have been indirectly induced. Intuitively, there can be no causal influence if there is no structural connection (either direct or indirect) for one node to influence the other. An example is shown in \textbf{B} for node $i$ that does not receive any input from other nodes and, consequently, its activity is not impacted by any perturbations, which is not the case for node $j$.}
\label{fig2}
\end{figure*}

\subsubsection{Susceptible-Excited-Refractory Model}\label{SER}
SER provides an abstract model of a spiking neuron and has been previously exploited to investigate the propagation of excitation in networks of interconnected nodes with three discrete states, $\{S, E, R\}$\cite{Garcia2012-ir,Messe2015-lh}. At each time step, $t$, the state of each node is determined based on the following rules:
\begin{itemize}
\item Nodes in $S$ state at step $t-1$ can enter the state $E$ at step $t$ either spontaneously with a probability of $p=0.005$ or if the weighted sum of the inputs passes a threshold $\theta=0.6$.
\item Nodes in state $E$ at time $t-1$ enter the state $R$ at time $t$.
\item And lastly, nodes in state $R$ at time $t-1$ switch back to $S$ at $t$ with a probability of $p=0.3$ or remain in state $R$ with a probability of $1-p$.
\end{itemize}

\begin{figure}[ht]
\centerline{\includegraphics[scale=0.9]{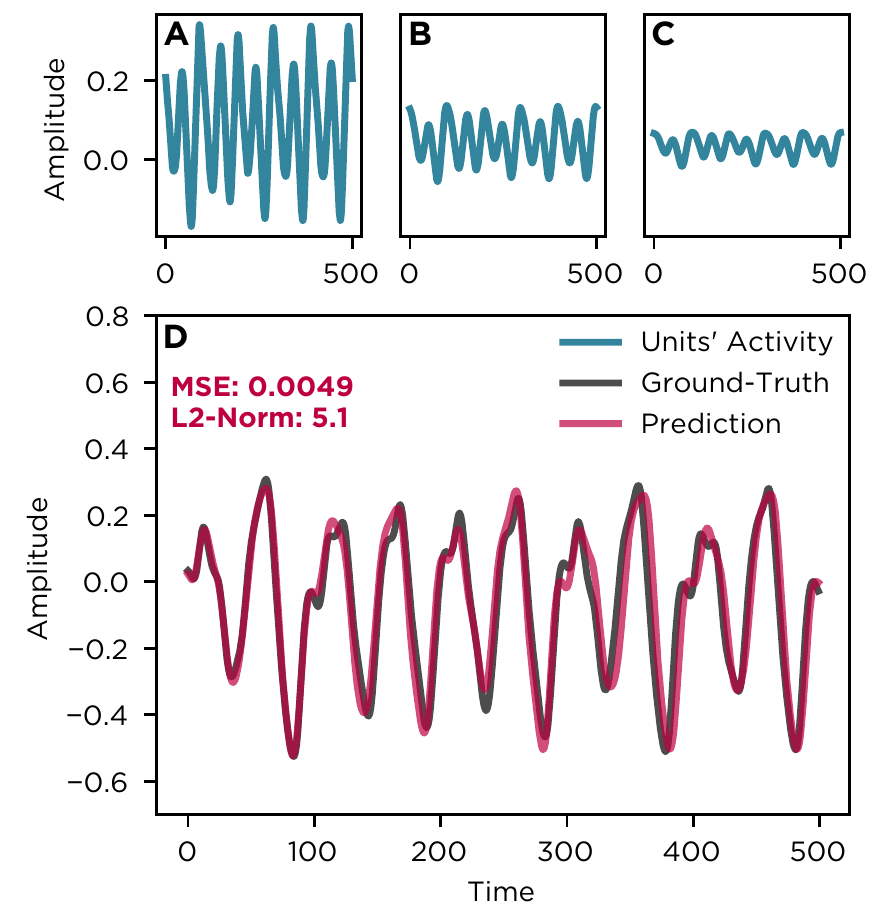}}
\caption{\textbf{Performance and internal representations of the ESN:} The three top panels \textbf{A, B,} and \textbf{C} show the internal representations of three random nodes 1, 18, and 21 after the ESN (with parameter set $\rho = 0.66$ and $a = 0.1$) is trained. Panel \textbf{D} shows the performance of the network (red) superimposed on the ground-truth signal (dark grey).}
\label{fig3}
\end{figure}

For the SER model, negative weights were not implemented since, by definition, nodes cannot suppress their neighbors. Thus, effectively, negative values for weights serve as `no connection'. To compute the causal influences in this network (see section: \nameref{CID}) we simulated the SER model from 5000 random initial conditions and recorded the sequence of discrete excited events from each node for 500 time steps. Each sequence was then convolved with a Gaussian kernel with $\sigma=10$ to transform the discrete-state time series into continuous-state sequences.

\subsection{Systematic Perturbation Framework}\label{Systematic Perturbation}
\begin{figure*}[!tb]
\centerline{\includegraphics[scale=0.95]{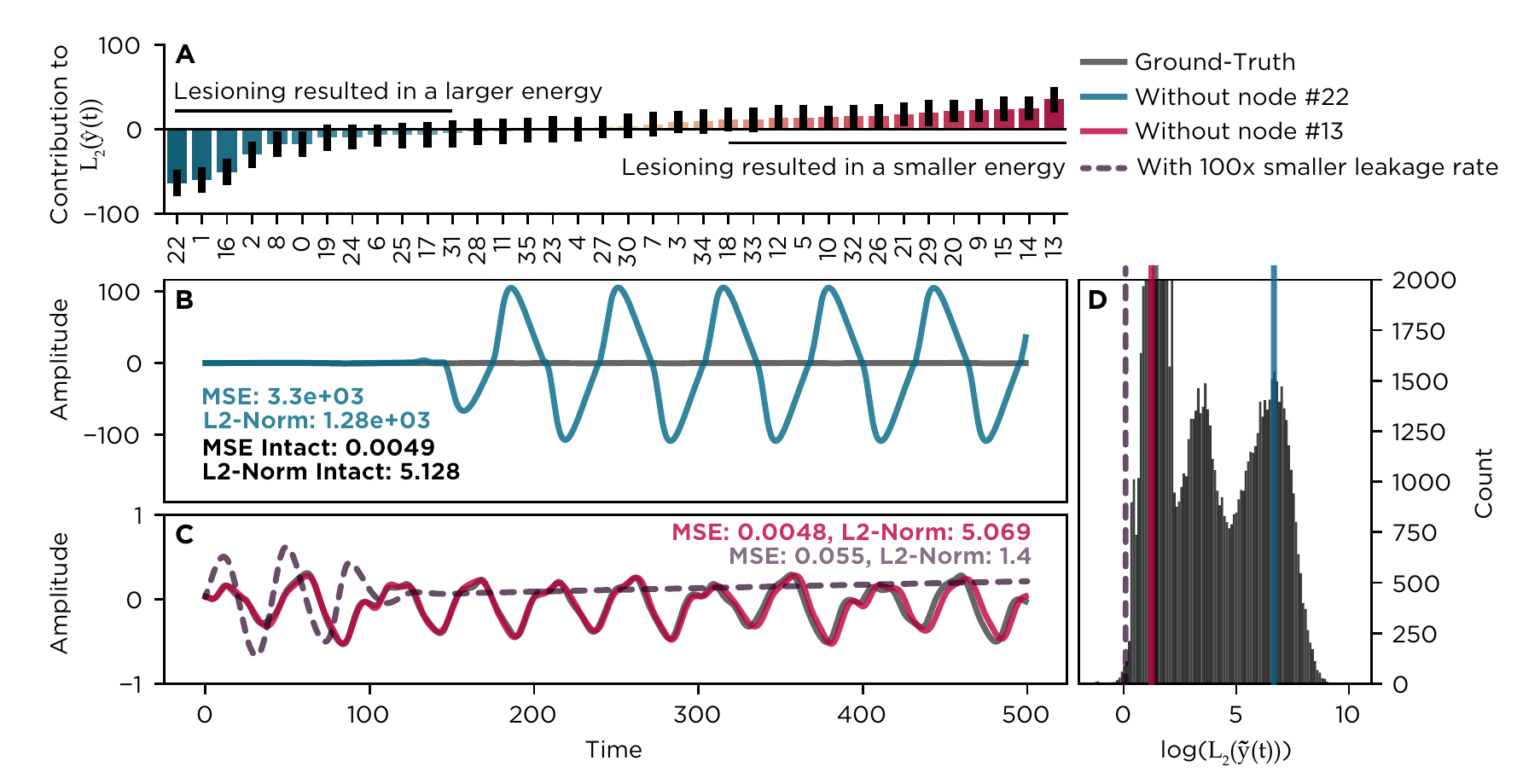}}
\caption{\textbf{Ranking nodes according to their contributions to behavior:} The \textbf{A} shows the contribution of each node to behavior, quantified by MSA, by tracking changes in $L_2(\tilde{y}(t))$. Black lines are 95\% confidence interval. Note that most nodes are involved to some degree and their contributions add up to $5.128$, which is the energy of the intact prediction, i.e., $L_2(\hat{y}(t))$. This means the behavioral performance is fairly distributed across its underlying involved nodes such that the larger the positive contribution is, the bigger share it has in producing $\hat{y}$ and the larger the negative contribution, the more critical role a node plays to keep $\hat{y}(t)$ from over-activation. This is confirmed in the two panels \textbf{B} and \textbf{C} in which removing the node \#13 resulted in a $L_2(\tilde{y}(t)) < L_2(\hat{y}(t))$ while removing node \#22 ended with an astronomically large $L_2(\tilde{y}(t))$. The histogram in the panel \textbf{D} shows the natural log of every produced $L_2(\tilde{y}(t))$ following a perturbation. The total number of perturbation is around 360.000 and the y-axis is clipped for clarity. The largest bin contains 12.700 perturbations.}
\label{fig4}
\end{figure*}

In this work, we define the `cause' based on the manipulability theories of causation \cite{Woodward2009-hx}. Specifically,  ``an element, $i$, has a causal contribution to (or is causally influencing) another element, $j$, if manipulating $i$ leads to a tractable change in $j$". We use `perturbation' as our preferred method of manipulation. However, the choice of metric by which the effect of perturbation is characterized directly impacts the findings. For instance, perturbing $i$ might change $ j$'s baseline activity while only the change in its variance is being tracked. Thus, to be sensitive to both, we tracked the change in the \textit{energy} of the signal defined as the $p$-th Euclidean norm of a continuous-state sequence, $\tilde{x}(t)$:

\begin{equation}
\label{eq: eq3}
L_{p}(\tilde{x}(t))=\left(\int_{-\infty}^{\infty}|\tilde{x}(t)|^{p} d t\right)^{1 / p},
\end{equation}

with $p=2$. For instance, to calculate the contribution of nodes to behavior, we initially computed the energy of the predicted output, $\hat{y}(t)$, of the intact network to establish a baseline (Fig.\ref{fig4}). Perturbing a node that has no causal contribution to the signal produced by the network, would, therefore, result in $L_{2}(\hat{y}(t)) = L_{2}(\tilde{y}(t))$ where $\tilde{y}(t)$ indicates the output of the \textit{perturbed} network. In contrast, removing a set of crucial nodes may push the system towards an equilibrium where $\tilde{y}(t) = \emptyset$ and $L_{2}(\tilde{y}(t)) = 0$. Another extreme behavior occurs when perturbing a different set of nodes results in a high amplitude oscillatory pattern in the predicted output and, consequently, a tremendous increase in the signal's energy. In both cases, perturbed sets are causally contributing to the performance by increasing or decreasing the signal's energy.

\subsubsection{Multi-perturbation Shapley value Analysis}\label{MSA}
To uncover the causal role of brain regions, conventionally, a single element is perturbed, and the consequent effect is measured. Cognitive functions, however, emerge from higher-order interactions of many elements \cite{Misic2016-as}. These interactions render the results from perturbation of single elements misleading \cite{Fakhar2021-jd,Jonas2017-gx}. For instance, Fig\ref{fig2}.B illustrates a scenario in which removing a pair of nodes \{D, A\} (the blue time-series) provides a categorically different result than lesioning one of the elements, e.g., \{D\} (the red time-series).

To account for these effects we used multi-perturbation Shapley value analysis (MSA) \cite{Keinan2006-kl,Keinan2004-om} that is based on computing the so-called Shapley value, $\gamma$, an axiomatic game-theoretical solution for fairly allocating the outcome of a coalition to its members \cite{Shapley1953-us}. $\gamma_i(N,\nu)$ can be interpreted as the `worth' of $i$ for the grand coalition $N$ based on a value function, $\nu$. Here, $\nu$ is the energy of the $\hat{y}(t)$ for the output signal and $x(t)$ for individual nodes. The worth of elements is then calculated by their contributions to the quantified energy. Further, $\gamma_i(N,\nu)$ is derived from all possible combinations in which $N$ forms groups with and without $i$. Formally, the marginal contribution of $i$ to a set $S$ with $i\notin S$ is:

\begin{equation}
\Delta_{i}(S)=\nu(S \cup\{i\})-\nu(S).
\end{equation}
Then $\gamma_i$ for $i \in N$ is defined as:
\begin{equation}
\gamma_{i}(N, \nu)=\frac{1}{n !} \sum_{R \in \mathcal{R}} \Delta_{i}\left(S_{i}(R)\right),
\end{equation}

Where $\mathcal{R}$ is the set of all $n!$ orderings of the grand coalition $N$ and $S(R)$ is the set of players preceding $i$ in the ordering $R$. Since reaching an analytical solution to $\gamma$ for large sets could be computationally prohibitive, we used an unbiased estimator that samples coalitions from the space of $2^M$ possible combinations, where $M$ is the number of all elements in the grand coalition \cite{Keinan2006-kl}. In our ESN and SER experiments, 5000 permutations per element were performed to infer the causal influence of nodes on each other, and 10.000 permutations were used to compute the contribution of ESN nodes to its performance (\textit{i.e.}, the predicted output, $\hat{y}(t)$). These numbers are 1000 and 10.000 for the control ESN, respectively.

To quantify the contribution of each node to $\hat{y}(t)$, we performed the MSA on all nodes while tracking the $L_{2}(\tilde{y}(t))$. For every permutation, first, a set of nodes excluding a target node is perturbed by setting their connection weights to zero, an $L_{2}(\tilde{y}(t))$ is computed, the target node is then perturbed alongside the lesioned set, and the $L_{2}(\tilde{y}(t))$ is computed again to obtain $\Delta_{i}(S) = (L_{2}(\tilde{y})^{i \in S}(t) - L_{2}(\tilde{y})^{i \notin S}(t))$, \textit{i.e.,} the contribution of the target node $i$ to that specific set of elements, $S$.

For example, if the contribution of a set $S=\{i,j,k\}$ to the energy of the output signal were $L_{2}(\tilde{y}(t))=5$ and $S=\{i,j\}$ resulted in $L_{2}(\tilde{y}(t))=1$ then $\Delta_k=5-1=4$. This value can be interpreted as the added value of element $k$ to the set $\{i,j\}$ in producing $L_{2}(\tilde{y}(t))$ and not the contribution of $k$ to $L_{2}(\hat{y}(t))$, itself. Averaging over all permutations then results in the average marginal contribution of an element to all formed coalitions and, therefore, to producing the behavior $L_{2}(\hat{y}(t))$ (see Fig. \ref{fig2}.A for a schematic illustration).

\subsubsection{Causal Influence Decomposition}\label{CID}
Causal influence decomposition builds on MSA to quantify how much each node is influenced by the other nodes in a network. To elaborate, for each node, causal influence decomposition performs MSA on $N-1$ nodes while keeping track of the changes in the energy of the target node as $L_{2}(x_i(t))$. Similar to MSA itself, the goal is to decompose the causal contributions of each element. However, here the focus is shifted from the behavior of the network (i.e., predicted output, $\hat{y}(t)$) to the response of individual nodes $x(t)$. In this case, the most causally influential nodes for the target node $i$ are the ones that, if perturbed, would produce a large deviation in $L_{2}(x_i(t))$ both by silencing it and by causing over-activation. The least contributing nodes to $i$ are the ones with little influence over its dynamics, \textit{i.e.,} $L_{2}(x_i(t)) = L_{2}(\tilde x_i(t))$ where $L_{2}(\tilde x_i(t))$ represents the energy of the signal after perturbation (Fig \ref{fig2}.B). Note that causal influence decomposition captures indirect causal influences since removing an arbitrary node $j$ with no connection to $i$ can still impact its activity via perturbing nodes that are causally influencing the node $i$ (Fig.\ref{fig2}.C).

\section{Results and Discussion}\label{Results}

We implemented ESN and SER networks with `Echos' and `SER' python packages publicly available in \cite{Damicelli2019-wt,Damicelli2019-zd}. To perform MSA and causal influence decomposition, we utilized the Python package provided by \cite{Fakhar2021-jd}. The code for reproducing the experiments can be found in the following Github repository: \url{https://bit.ly/ESN_causal_influences}.

The MSE between $\hat{y}(t)$ and $y(t)$ shows that, after training, the ESN model could perform reasonably well in predicting the next 500 samples of the Mackey-Glass time series. Inspecting three exemplar nodes also shows that even though the network is rather compact with a densely clustered connectome, the internal representations were sufficiently diverse (Fig\ref{fig3}).

\subsection{Estimating Causal Contributions of Nodes to  Performance}\label{Results MSA}
After a systematic multi-site perturbation of all nodes, MSA ranked them such that those with a positive contribution ($\gamma_{i}>0$) are the ones that, when perturbed, on average, caused a reduction in the energy of the predicted signal such that $ L_{2}(\tilde{y}(t)) < L_{2}(\hat{y}(t))$. Fig.\ref{fig4}.C depicts an example where omitting node \#13 reduced $L_{2}(\hat{y}(t)) = 5.12$  to $L_{2}(\tilde{y}(t)) = 5.06$, and, interestingly, resulted in a smaller MSE. Additionally, removing the node with the highest negative contribution ($\gamma_{i}<0$), i.e., \#22 caused a large disruption in behavior, increasing both MSE and $L_{2}(\tilde{y}(t))$ drastically. Note that the sum of the individual contributions equals the energy of the well-predicted output of the intact network, $L_{2}(\hat{y}(t))$. This implies that MSA has distributed the produced outcome across 36 nodes such that each node received its `fair share' given its contribution.

Importantly, in Fig.\ref{fig4}.A the positive contributions steadily increase while the negative ones show a less linear trend. This observation can be explained as follows:
while almost half of the network was involved in producing the behavior (positively contributing nodes), a smaller set of negatively contributing nodes secured the stability of the network. Interestingly, the smaller set itself was divided into two sub-groups, a larger population with smaller contribution values and a localized set of 3-4 members with large contributions, hence resulting in a non-linear trend in the negative contributions instead of a steady decrease. This mechanism might also explain observation of two peaks in larger $L_{2}(\tilde{y}(t))$ values of the perturbation histogram depicted in Fig.\ref{fig4}.D.
The perturbation histogram was constructed from the resulting $L_{2}(\tilde{y}(t))$ of approximately 360.000 coalitions adopted by MSA to rank nodes of the network based on their contributions to prediction. The fact that it is a multimodal distribution points towards a handful of states in which the network mostly settled after lesions. Note that the y-axis is clipped for clarity and the largest bin has around 12.700 counts.

Additionally, both $L_{2}(\hat{y}(t))$ and $L_{2}(\tilde{y}(t))$ where node \#13 was lesioned fell into the largest peak. In comparison, the activity produced by the control ESN with inadequate leakage rate resulted in a small $L_{2}(\tilde{y}(t))$ that is located on the left edge of the histogram.

Our results suggest that the ESN model distributed the computations across most of the nodes, specifically, two functionally distinct subsets of nodes. However, this regime of distributed computation might not occur in larger networks or simpler tasks that are solvable by a small group of neurons where the remainder of the nodes potentially produce redundant representations of the external input. This phenomenon can be investigated by systematically increasing the size of the reservoir to identify if there is a minimum number of nodes required to achieve a particular task and, if so, what the role of other nodes would be.

\subsection{Decomposing Causal Influences of Nodes on Other Nodes}\label{Results CID}
\begin{figure}[hbtp]
\centerline{\includegraphics[scale=1]{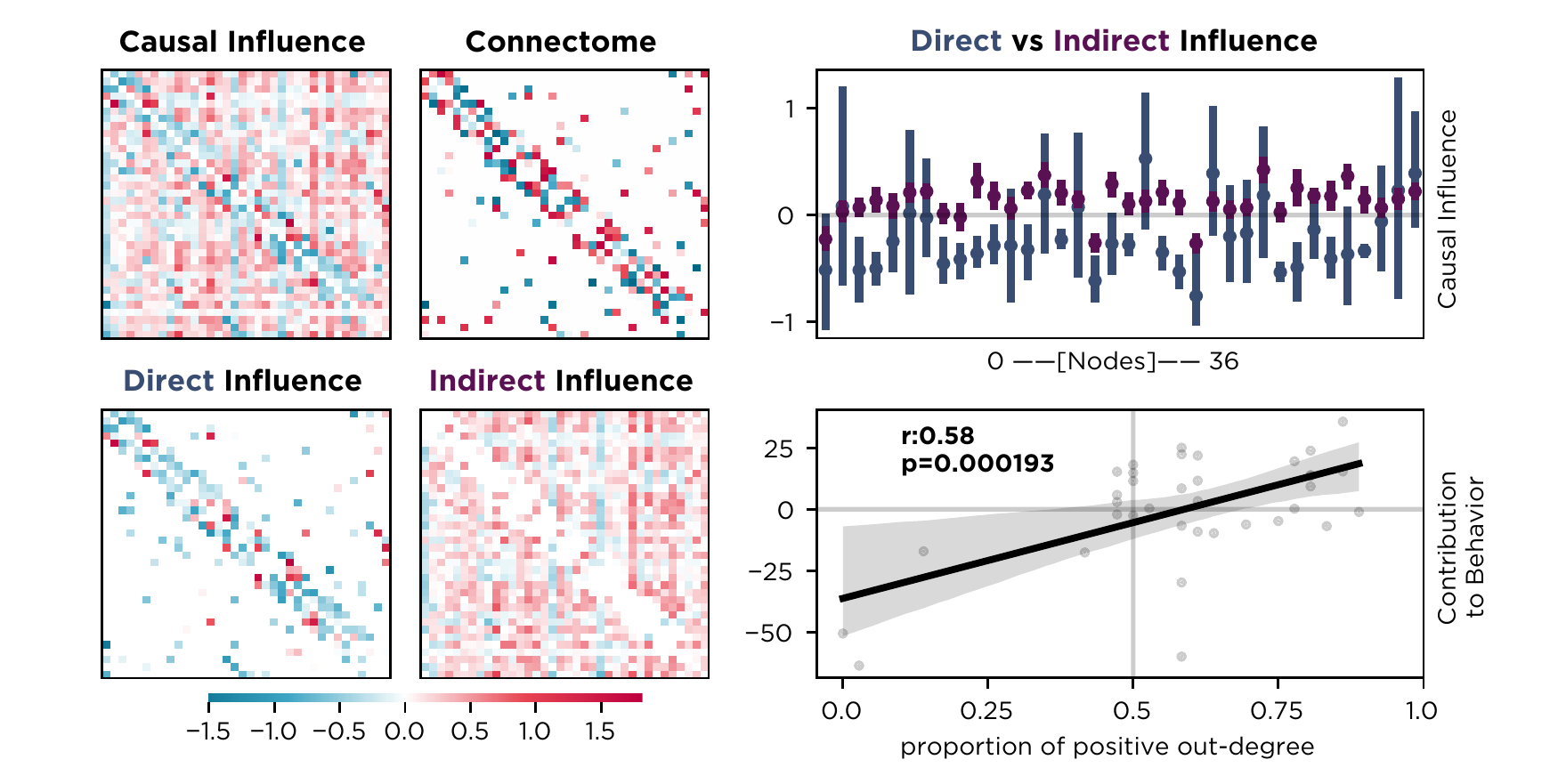}}
\caption{\textbf{Causal influence of nodes on each other:} The top two panels show the causal influence matrix and its underlying structural connectome (ESN's weights). A causal influence matrix consists of two additive components, direct and indirect causal influences.
Direct causal influences are the ones with a value in both matrices, while indirect ones are the ones only in the causal influence matrix. The top-right panel depicts the average causal influence of each node, divided into direct and indirect pathways. The error bar shows the 95\% confidence interval. The regression plot at the bottom-right shows the relationship between a node's influence and its contribution to the energy of the network's output. The gray area is the 95\% confidence interval.}
\label{fig5}
\end{figure}

After ranking the nodes according to their contributions to the performance, we computed their influence on each other. The result is shown in the causal influence matrix (Fig.\ref{fig5}) where each entity shows the influence of a node on the other, derived from the same perturbation pipeline, \textit{i.e.,} the systematic multi-site perturbation.
The structural connectome is also plotted to provide a reference for comparison. Given the structural connectome and the causal influence matrix, both direct causal influences \textit{i.e.}, the ones with a corresponding structural connection, and indirect causal influences \textit{i.e.,} the ones without the structural backbone, can be separately inferred (schematically shown in Fig.\ref{fig2}.C). It can be seen in Fig.\ref{fig5} that nodes largely influence each other through direct connections, while indirect influences were not completely obsolete. Interestingly, the majority of direct causal influences were negative, implying that removing them resulted in a destabilized target node with higher energy than its intact state (i.e., $L_{2}(\tilde{y}(t)) > L_{2}(\hat{y}(t))$). In contrast, indirect causal influences were mainly positive, meaning that while perturbing a node had a destabilizing effect on its immediate neighbors, the downstream effects of this destabilization reduced the energy of the distant nodes. 

Next, we ranked each node by its ratio of positive influences, \textit{i.e.,} how many of its interactions with other nodes were positive, meaning that they enhanced the energy of the target nodes. In this way, we assigned a small value to a node if lesioning it resulted in over-activation of most of the other nodes (see \textit{e.g.,} the vertical blue strips in the causal influence matrix depicted in Fig.\ref{fig5}). A large value then shows that the node had a large number of positive influences and removing it resulted in a lower energy of the other nodes. We found a correlation of Pearson’s $r = 0.58$, $p-value = 0.00019$ between this ranking and the one based on the contributions to the behavior (Fig.\ref{fig5}). This finding suggests that if lesioning a node had a destabilizing impact on the other nodes, it is more likely that the same node, when lesioned, disrupts the behavior of the whole network, too. Performing the causal influence decomposition resulted in two main findings:
\begin{enumerate}
\item A node's influence did not end with its connected neighbors, as each node received influences from all other nodes. This means perturbing a node had a global impact, with neighbors suffering the most.
\item Even though the direct influences were, on average, larger than the indirect ones, pointing towards a reliable structure-function relationship, this relationship was not straightforward.
\end{enumerate}

\begin{figure}[hbtp]
\centerline{\includegraphics[scale=1]{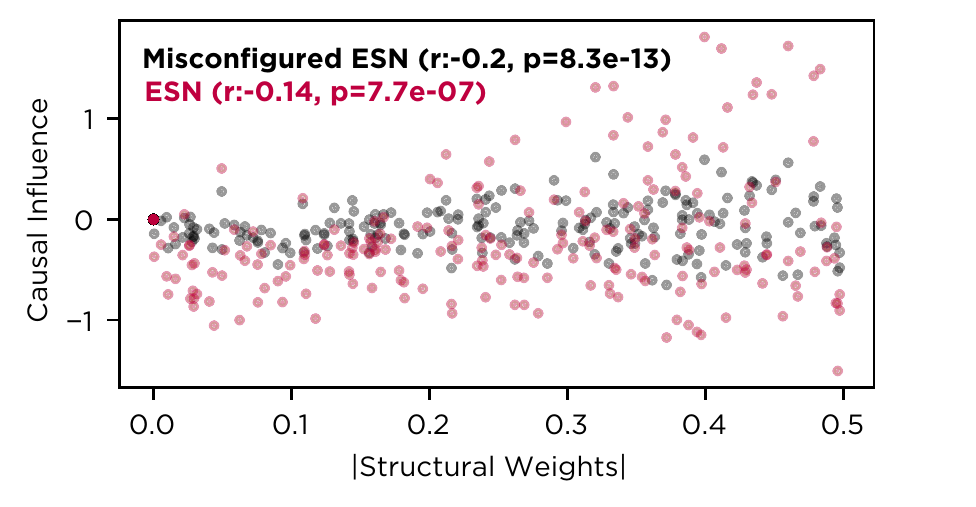}}
\caption{\textbf{Causal influences in well-designed and poorly-engineered ESNs:} This figure illustrates the scatter plot of absolute structural weights and causal influences in two properly and poorly engineered networks with the well-designed ESN represented in red and the control ESN in black. As depicted, the relationship is less linear with a smaller Pearson's correlation in the ESN than the poorly-engineered ESN.}
\label{fig6}
\end{figure}

To investigate if this decoupling is the key to a diverse and rich repertoire of internal representations in the reservoir (Fig.\ref{fig5}), we performed the same analytical pipeline of causal influence decomposition on both a control ESN with insufficient state representations as well as an SER model. As shown in Fig.\ref{fig6}, unlike the intact ESN, the relationship between causal influences and their structural connection weight tends to be linear in the poorly-engineered ESN. In fact, this relationship is strongest in the SER model (Pearson's $r = 0.64$, $p-value<0.0001$). This finding reveals a notable decoupling in both ESNs as opposed to the SER. In other words, the diverse internal representations produced by nodes in well-designed ESN were possibly less dictated by the connectome's weights.

To summarize, in comparison with the two above-mentioned control scenarios, the structural backbone played a less intuitive role in the interplay between nodes in ESN. In contrast, stronger underlying connections ensured more profound influences in the poorly-engineered ESN and the discrete excitable model. Interpreting these results, however, the following remarks should be taken into consideration.

The causal influences and contributions result from potentially thousands of perturbations to the system that were eventually reduced to a single value per element (Fig.\ref{fig2}.A). This means that the higher-order interactions are \textit{embedded} into these values but not further characterized. Additionally, although the contributions linearly add to the outcome of the grand coalition, which satisfies the third axiom of Shapley values defined in \cite{Shapley1953-us}, removing nodes does not necessarily impact the outcome proportionally. For instance, it is not guaranteed to have a reduction of $b$ points in the outcome following a lesion in $i$-{th} node with $\gamma_i(N,\nu)=b$. Again, the reason is that removing that element is a single case in thousands of combinations of lesions that were averaged to calculate the final value, $b$.

Next, causal influence decomposition does not `recover' the structural connectome. Given the connectome, our goal was to quantify how much each node was causally contributing to both the behavior and other nodes' dynamics. As shown here, some nodes displayed small contributions to others even though the underlying structural link between them was strong. Others had a major impact on the rest of the network, even with a weak structural connection. Indirect contributions added to this complexity by demonstrating that a node could still influence others without any structural connection, albeit not as strongly.

Moreover, we revealed functionally distinct units with positive and negative contributions. Combining this finding with the decoupling of structure-causal influences suggests that structurally excitatory and inhibitory connections do not necessarily excite or inhibit their targets. In fact, in our ESN, structurally positive weights were mainly preventing their targets from destabilizing (Fig\ref{fig5}).

Additionally, here we used an ESN with rate units. Discrete-state nodes such as spiking neural networks can be included to comprehensively compare structure-function and behavior in both continuous- and discrete-state models. The only limitation, which can be prohibitive, is the computational cost of the methods.

To conclude, we introduced a novel approach called causal influence decomposition based on the multi-perturbation Shapley value analysis. Both frameworks rely on systematic multi-site perturbation as the mean for revealing causation in the system, given a cause being a contributor to its effects. MSA quantifies the causal contribution of elements to the outcome of the system  which they collectively construct. Causal influence decomposition quantifies the influence of elements on each other. We emphasize the `systematic' and` multi-site aspects of the perturbations since, as shown, neglecting either of these even in small systems \cite{Fakhar2021-pg} or performed exhaustively \cite{Jonas2017-gx} results in misleading findings. Moreover, we advocate \textit{in-silico} experiments on ground-truth models such as ESNs to verify fundamental assumptions of the employed methods in neuroscience. ESNs are particularly suitable for this task since the mapping between structure, function, and behavior can be investigated extensively.

\section{Acknowledgments}\label{Acknowledgments}
Funding is gratefully acknowledged: KF: Deutsche Forschungsgemeinschaft, Germany (178316478-A1; TRR169/A2). 
FH: Deutsche
Forschungsgemeinschaft, Germany (178316478-A1; TRR169/A2). 
CCH: Deutsche Forschungsgemeinschaft, Germany (178316478-A1; TRR169/A2; SFB 1461/A4; SPP 2041/HI 1286/7-1, HI 1286/6-1), the Human Brain Project, EU (SGA2, SGA3).

\bibliographystyle{unsrt}  
\bibliography{references}

\end{document}